# Unveiling Glitches: A Deep Dive into Image Encoding Bugs within CLIP

**Ayush Ranjan, Daniel Wen, Karthik K Bhat**
aranjan1@ucsc.edu, dywen@ucsc.edu, kabhat@ucsc.edu

## Abstract

Understanding the limitations and weaknesses of state-of-the-art models in artificial intelligence is crucial for their improvement and responsible application. In this research, we focus on CLIP, a model renowned for its integration of vision and language processing. Our objective is to uncover recurring problems and blind spots in CLIP's image comprehension. By delving into both the commonalities and disparities between CLIP and human image understanding, we augment our comprehension of these models' capabilities. Through our analysis, we reveal significant discrepancies in CLIP's interpretation of images compared to human perception, shedding light on areas requiring improvement. Our methodologies, the **Discrepancy Analysis Framework (DAF)** and the **Transformative Caption Analysis for CLIP (TCAC)**, enable a comprehensive evaluation of CLIP's performance. We identify **14 systemic faults**, including Action vs. Stillness confusion, Failure to identify the direction of movement or positioning of objects in the image, Hallucination of Water-like Features, Misattribution of Geographic Contex, among others. By addressing these limitations, we lay the groundwork for the development of more accurate and nuanced image embedding models, contributing to advancements in artificial intelligence.

## 1 Introduction

In recent years, the integration of vision and language processing has propelled the field of artificial intelligence towards new horizons, enabling models like CLIP to achieve remarkable feats in image comprehension. However, as with any tool, CLIP has its limitations and areas where it struggles. While prior research has predominantly concentrated on evaluating the text encoder functionality of CLIP, as demonstrated in the paper(1), we argue that exploring the image embedding aspect of CLIP is just as important and valuable as investigating its other features. By scrutinizing both components, we can achieve a more comprehensive comprehension of CLIP's strengths and weaknesses.

This paper delves into uncovering the recurring problems and blind spots in CLIP's image comprehension. By scrutinizing both the commonalities and disparities between CLIP and human image understanding, we aim to augment our comprehension of these models' capabilities. To achieve this, we employ two novel methodologies: the Discrepancy Analysis Framework (DAF) and the Transformative Caption Analysis for CLIP (TCAC).

The **DAF** method facilitates a systematic examination of CLIP's performance by analyzing discrepancies in image similarity rankings between CLIP and DINOv2(12), while the TCAC approach enables a comprehensive evaluation of CLIP's response to transformations applied to images. Through these rigorous methodologies and analysis, we aim to identify systematic failures in CLIP's interpretation of images, shedding light on specific areas where improvement is needed.

Understanding the limitations of CLIP holds paramount importance for several reasons. Firstly, it unveils potential biases within its training data and architecture, laying the foundation for targeted enhancements. Secondly, it serves as a guide for the responsible application of CLIP, pinpointing areas where accuracy or completeness may be compromised. Lastly, these insights pave the way for the development of more comprehensive and nuanced image embedding models, propelling progress in the broader landscape of artificial intelligence.

By focusing on CLIP's image comprehension, we contribute to bridging the gap in existing research and provide valuable insights for improving multimodal models. Through rigorous methodologies and analysis, we aim to uncover novel perspectives on CLIP's weaknesses, paving the way for advancements in artificial intelligence and multimodal understanding.

## 2   Related Work

**Evaluation of multimodal models:** Existing research (1) tackles failures in NLP (factual consistency, bias) and CV (adversarial examples) often through manual curation or specific failure modes. Multimodal evaluation focuses on output quality, not systematic errors. The paper addresses the gap by introducing MultiMon, which automatically identifies systematic failures in multimodal systems with language models. MultiMon leverages language models to uncover generalizable error patterns, offering a more scalable and comprehensive evaluation approach.

**Text-guided multimodal models:** In this work, we analyze the shortcomings of text-guided multimodal generation models. These models can create various outputs, including images, videos, and 3D scenes, based on textual descriptions (2). Commonly, these models employ a vision-language model (VLM) to encode the text input. VLMs establish a shared embedding space for both text and image data (3). Subsequently, a guided diffusion process is used to generate the final output (4).

**Vulnerability of vision and language models:** Several studies (5) have explored robustness in vision and language models. For instance, prior work demonstrated that Transformers trained with Masked Language Modeling exhibit limited sensitivity to word order (6). This suggests BERT's success might rely more on co-occurrence statistics than true syntactic and semantic understanding (7). Existing benchmarks evaluate robustness against image corruptions, style changes, and viewpoints, primarily focusing on unimodal tasks (vision-only or language-only) (8). Our work differs by investigating CAB, a cross-modal phenomenon requiring both image and language data. While studies have explored compositional generalization and probed visuo-linguistic abilities with datasets like Winoground, models like CLIP (9) struggle on these tasks. This highlights the need for models to go beyond basic compositional understanding and incorporate reasoning and object detection. Our work complements these findings by revealing CLIP's brittleness to CAB, a previously unexplored vulnerability.

**Peculiarities of CLIP:** Recent works highlight limitations in the compositional reasoning abilities of state-of-the-art image generation models, including DALL-E 2 (9). CLIP-based image encoders are suspected to contribute to these shortcomings (10). Notably, models like Imagen and Parti, which lack CLIP, demonstrate superior performance in tasks requiring compositional reasoning (11). However, a deeper understanding of CLIP's behavior in tasks like zero-shot image classification and visual question answering is lacking. Our work bridges this gap by analyzing CLIP through the lens of CAB, revealing a novel perspective on its weaknesses in compositional reasoning.

## 3   Approach

In this study, we employed two novel methodologies: the Discrepancy Analysis Framework (DAF) and the Transformative Caption Analysis for CLIP (TCAC). The TCAC approach can be viewed as an extension of the DAF method, facilitating a comprehensive evaluation of our research outcomes.

## 3.1 Methodology 1: Discrepancy Analysis Framework (DAF)

### 3.1.1 Step 1: Data Acquisition and Preprocessing

Our analysis commenced with the utilization of the Flickr8k dataset, known for its extensive collection of images matched with multiple captions. To streamline our procedures, we systematically pre-processed the dataset, ensuring a reliable association between image filenames and their respective captions. Initially, we opted for the first caption associated with each image from the Flickr8k dataset. However, we noted a deficiency in the descriptive quality of these captions regarding the image content. Subsequently, after assessing numerous images, we determined that the longest caption consistently offered the most comprehensive information pertaining to the image. As a result, we transitioned to utilizing the longest caption for our analytical endeavors.

### 3.1.2 Step 2: Image Embedding Extraction

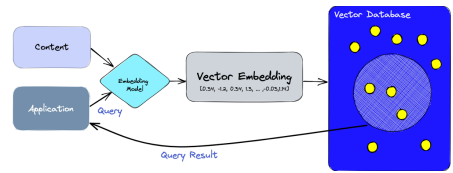

Figure 1: Image Embedding Extraction

Utilizing the capabilities of both CLIP and DINOv2 (12) models, we proceeded to extract image embeddings for every image within our dataset. DINOv2 proves to be a robust option for extracting image embeddings in our scenario due to its notable advantages. It outperforms CLIP notably in tasks related to image similarity, ensuring higher accuracy. Additionally, DINOv2 is trained exclusively on visual data, resulting in a cleaner representation focused solely on the content of the images.

These embeddings were subsequently stored in specialized vector databases (13), guaranteeing streamlined retrieval and comparison during subsequent stages of analysis.

### 3.1.3 Step 3: Comparative Examination of CLIP and DINOv2 Similarity Rankings

We employed mathematical methodologies, such as L2-similarity computations, as an initial step. Subsequently, we retrieved the top 10 most similar images based on the embeddings generated by CLIP and DINOv2, respectively. We identified instances of "failure" cases, characterized by divergent similarity assessments between the models. Based on insights gleaned from multiple research papers, we transitioned to utilizing cosine similarity instead of L2, as it was determined to yield more accurate results.

### 3.1.4 Step 4: Identifying Failure Cases using GPT

We utilized the sophisticated language processing capabilities of GPT-3.5 to conduct a comprehensive analysis of the captions. The discrepancies in captions were inputted into ChatGPT using a specific prompt (shown in 2), aimed at identifying significant issues.

### 3.1.5 Limitations of DAF

**Ensuring Dataset Consistency:** The accuracy of our analysis hinges significantly on the dataset's consistency, particularly concerning image similarity. A lack of diversity or inadequate coverage of various scenarios within the dataset may compromise the authenticity of the identified model failures.

**Importance of Descriptive Captions:** The quality and level of detail in captions are pivotal factors in discerning model failures. Vague or insufficiently descriptive captions may lead to inaccuracies in identifying similarities, potentially misrepresenting the limitations of the model.

> I will provide a series of data for you to remember. Please memorize this data because I will ask you questions on them afterwards:
>
> Data format: The first sentence is the main caption of the image, and the subsequent sentences are the captions generated by an embedding model on the same image.
>
> Task: I am trying to find failures with an embedding model. The above are sentences of image captions that it encodes very similarly, even though they are conveying different concepts. Using these specific examples, are there any general types of failures you notice the embedding is making, or any common features that the embedding fails to encode? Try to give failures that are specific enough that someone could reliably produce examples that the embedding would encode similarly, even though it shouldn't. Please try to give as many general failures as possible, and please just state each failure without further explanation. Please focus on differences that are important visually, as these embeddings are later used to generate images or videos.
>
> Here is an example for the format: {{"'Failure Type Heading'": "'Description: Describe in few words as to why the failure occurred'"}}
>
> Give the output as Python Dictionary of objects

Figure 2: GPT prompt

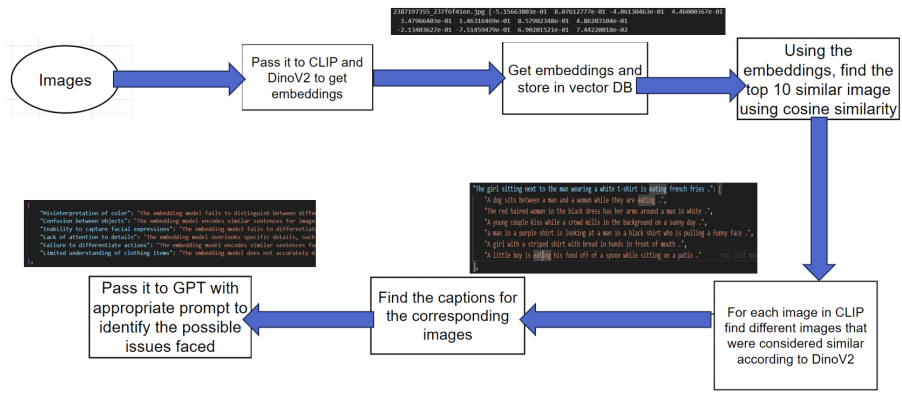

Figure 3: Overview of DAF

### 3.2 Methodology 2: Transformative Caption Analysis for CLIP (TCAC)

#### 3.2.1 Step 1: Transformation Setup

Creating a thorough range of image transformations to mimic real-world variations is crucial for enhancing the dataset. These transformations encompass various adjustments such as converting images to grayscale, implementing random rotations, and employing random affine transformations, among others. Each transformation introduces subtle changes to the images, mirroring the diverse conditions found in real-world scenarios and consequently influencing their encoding. Our goal was to select transformations that preserve the original information of the images without distorting them. After experimenting with several options, we settled on six transformations (show in 4) that meet this criterion.

#### 3.2.2 Step 2: Dataset and Preprocessing

Utilized the Flickr8k dataset, containing 5 captions per image, and applied CLIP text embedding to identify similar captions for each original caption. Generated a list of 50 captions by finding 10 similar captions for each original caption and removed duplicates to obtain a list of unique captions.

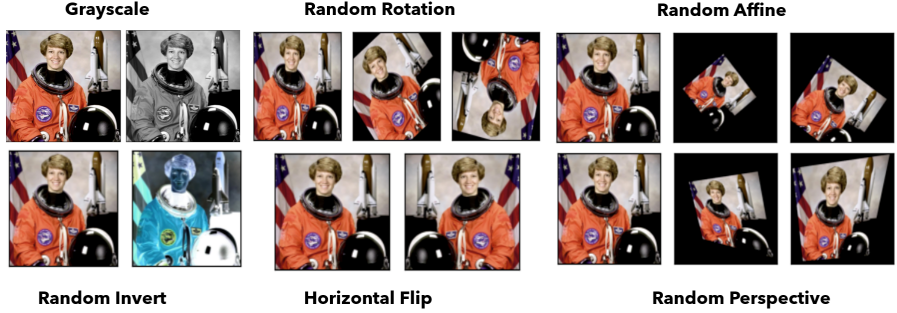

Figure 4: Transformation Setup

### 3.2.3 Step 3: Transformation Probability Prediction

After applying these transformations, we utilize CLIP's advanced capabilities to predict the probability distribution across the set of image captions associated with each transformed version. This methodical approach allows us to thoroughly investigate the model's capacity to recognize and interpret visual content across a range of altered representations, thereby enhancing our understanding of its performance in varied conditions.

We predict the probability of each unique caption for every image, identifying the top 10 captions before and after transformation. Subsequently, we manually inspect images and captions for evaluation. To streamline the process, we develop a custom class that contains the image name, original top 10 captions, transformed top 10 captions, and the difference in caption count. We then sort these classes based on the difference in caption count and analyze 100 such classes for each type of transformation.

### 3.2.4 Step 4: Evaluation Process

In this step, we begin by manually inspecting the classes created in the previous step, analyzing both the captions before and after transformation. We carefully compare the original top 10 captions with the transformed top 10 captions to assess any discrepancies or changes induced by the applied transformations.

Additionally, we consider errors identified by the DAF method (if applicable). These errors provide valuable insights into potential inconsistencies or inaccuracies in the predicted captions. To verify the presence of errors, we manually examine how similar the original captions corresponding to the image are to those identified by the DAF method.

By combining manual inspection with the analysis of DAF errors, we gain a comprehensive understanding of the impact of transformations on caption prediction consistency. This thorough evaluation process allows us to identify transformations that introduce significant alterations in predicted caption probabilities and helps refine our approach accordingly.

## 4 Results

### 4.1 Raw Images

These are the images extracted from the Flickr8k dataset with no transformations done to them.

Figure 5 shows three examples of the systemic faults that were found in CLIP's image embedding capabilities (Section 4.3).

Image 5a is an example of CLIP's **misinterpretation of color**. It predicts the caption, "A girl in orange shirt and clown makeup stands in a park while others look on", to be a more accurate description of the image than the caption, "A man with orange boots stands with a few other people."

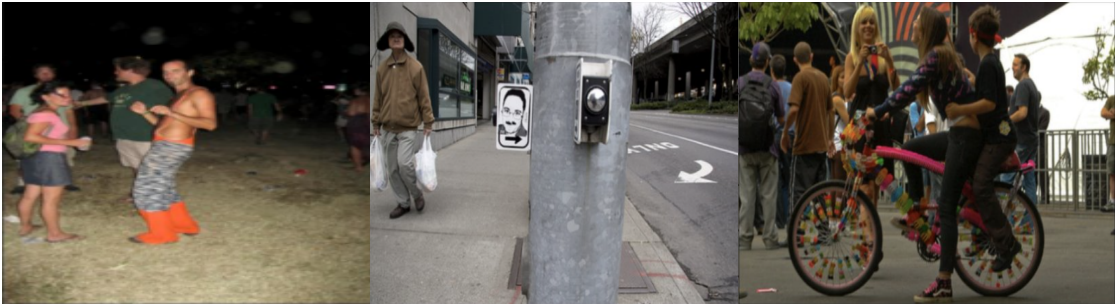

(a) Image 1  (b) Image 2  (c) Image 3

Figure 5

We observe that CLIP mistakes the girl's shirt to be orange due to the emphasis placed on the man's boots.

Image 5b is an example of CLIP **disregarding object interactions**. It predicts the caption, "A man is walking down the street holding up a picture", to be a more accurate description of the image than the caption, "A man carries two grocery bags, a sign of a funny man is by the pole near him." We see that CLIP understands that there is a man and a picture; however, it fails to recognize that the sign interacts with a pole instead of the man.

Image 5c is an example of how CLIP **confuses different modes of transportation**. It predicts the caption, "A woman wearing a pink sweater rides a bike with a car behind her", to be a more accurate description of the image than the caption, "A brightly decorated bicycle with cart with people walking around in the background." While CLIP correctly identifies the bicycle, it hallucinates a car behind it. Notice how the latter caption still hallucinates a cart attached to the bicycle; however, this is a more reasonable hallucination than the car, because the metal fence in the background appears similar to a cart/cage.

### 4.2 Transformed Images

These are the images after the transformations from Methodology 2 are applied to them.

Image 6a and 6d are examples of CLIP's **lack of attention to details** and **inability to capture age**. For 6a, it predicts the caption, "The tennis player holds the tennis racket in preparation for the ball", to be a more accurate description of the image than the caption, "A boy holds a net with a snake in it close to the ground." After the random inversion transformation, which negates pixel colors in the image at random, CLIP could no longer identify the snake and incorrectly identified the age of the boy. For 6d, it predicts the caption, "A man in a yellow hi-viz jacket is standing beside an orange sign", to be a more accurate description of the image than the caption, "A man holds a sign reading 'HOME ANYTHING HELPS' near a construction site." We observe this loss in detail of the sign to be somewhat reasonable as the words are flipped backwards after the horizontal transformation; however, CLIP thinks the sign is actually is a yellow hi-viz jacket instead.

Image 6b is an example of CLIP's **failure to properly encode different activities and positions of people**. It predicts the caption, "A man is helping a girl sit on a large bicycle" to be a more accurate description of the image than the caption, "A girl sits on a decorated bike with a younger boy while another girl takes a picture." While the first caption is not unreasonable, it lacks the details that the latter caption has; furthermore, the latter caption was not even in the top five captions after the grayscale transformation, while it was the top ranked caption of the raw image.

Image 6c is an example of CLIP's **failure to account for the size and number of objects/persons/animals** present in the image. It predicts the caption, "Three children wearing soccer uniforms chase after a blue and grey soccer ball" to be a more accurate description of the image than the caption, "Three people playing jump rope in a field." While the latter caption is less descriptive, it correctly identifies the activity of jump rope unlike the first caption, which thinks that the children are playing soccer. CLIP might understand that the children are playing a game, but it is unable to categorize the jump rope.

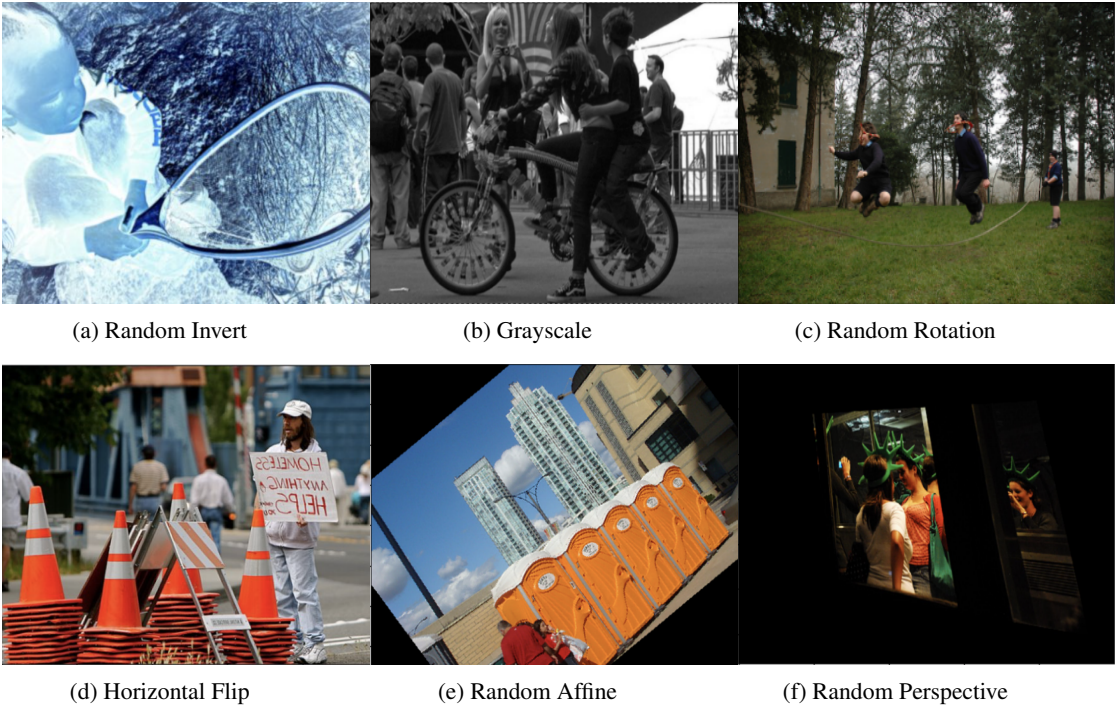

(a) Random Invert  (b) Grayscale  (c) Random Rotation

(d) Horizontal Flip  (e) Random Affine  (f) Random Perspective

Figure 6

Image 6e is an example of CLIP's **confusion between objects**. It predicts the caption, "A man walking in a brush with a big orange backpack", to be a more accurate description of the image than the caption, "A family waits next to some portable toilets in the city." The latter caption is present before the transformation, while the first caption is after. After the transformation, CLIP confuses the orange portable toilets with a big orange backpack. Random affine moves the image horizontally and/or vertically, rotates, resizes, and shears the image by a random amount, which alters the image more than the other transformations tested. This can be a contributing factor to the large erroneous caption after the transformation.

Image 6f is an example of CLIP's **misattribution of geographic context**. It predicts the caption, "Some young people take photos of each other while in New York", to be a more accurate description of the image than the caption, "Three women with statue of liberty hats and taking a picture." There is no way to assume that the image takes place in New York, but the statue of liberty hats confuses CLIP into believing so. However, before the random perspective transformation, CLIP was able to successfully identify the hats as costumes instead of assuming the women are in New York.

Image 7a is an example of CLIP's **hallucination of water-like features**. It predicts the caption, "Child, man, and woman walking near the water", to be a more accurate description of the image than the caption, "Several people are walking down a trial." While the latter caption is not the most descriptive, the first caption hallucinates a body of water when the image has none present. This is also common in elastic transformations due to the wavy features that inhibit CLIP's image embeddings of background sceneries.

Image 7b is an example of CLIP's **failure to differentiate action vs. stillness**. It predicts the caption, "A woman and boy playing on a beach", to be a more accurate description of the image than the caption, "The boy and girl are standing in front of the ocean and mountains." Not only do we observe loss of background details, CLIP miscategorizes the action of the kids as "playing" instead of "standing". Most of the top ten captions before the transformation say that the kids are either posing for the picture or standing still, while the top ten captions after the transformation say they are playing in the sand.

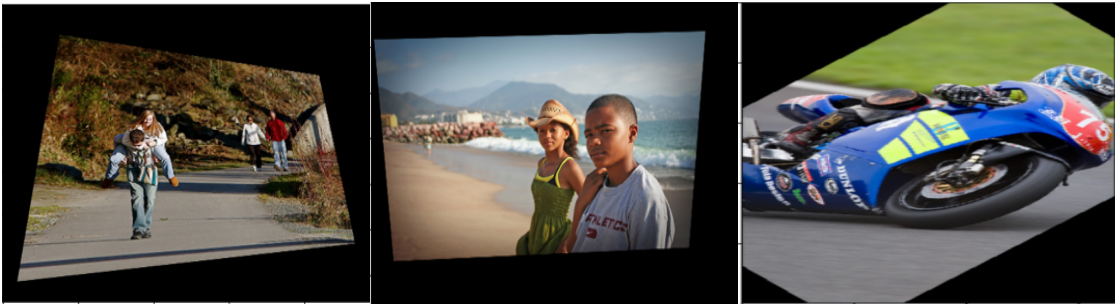

(a) Random Perspective        (b) Random Perspective        (c) Random Rotation

Figure 7

Image 7c is an example of CLIP's **failure to identify the direction of movement or positioning of objects in the image**. It predicts the caption, "A motorcycle racer leaning over towards the right to take a sharp turn", to be a more accurate description of the image than the caption, "A competitive motorcycle racer prepares to make a left turn along a paved road." While directional confusion can be understandable for transformations such as horizontal flip but for random rotation it shouldn't, CLIP still has trouble understanding the direction of movement based off the positioning of objects even when the transformation does not drastically alter the image.

### 4.3 Systematic Faults

Here are descriptions for the 14 errors we identified in CLIP's image embedding capabilities. The **bold systemic fault are novel faults** found in CLIP's image embedding capabilities. The non-bold systemic failures are known vision model faults from CLIP and other models found from other papers (14).

- **Action vs. Stillness confusion** - CLIP may struggle to distinguish between images depicting action or movement and those showing static scenes, leading to misinterpretations in image understanding.
- **Failure to identify the direction of movement or positioning of objects in the image** - CLIP may fail to accurately determine the direction of movement or the spatial arrangement of objects within an image, resulting in misinterpretations of scene dynamics.
- **Hallucination of Water-like Features**- CLIP may erroneously perceive water-like features in images where they do not exist, potentially leading to misinterpretations of the depicted environment.
- **Misattribution of Geographic Context**- CLIP may incorrectly attribute geographic context to images, such as misidentifying landmarks or landscapes, which could impact the model's understanding of the depicted scene.
- Misinterpretation of color - CLIP may inaccurately perceive or interpret colors in images, leading to errors in understanding the visual content.
- Confusion between objects - CLIP may struggle to differentiate between similar objects in images, resulting in confusion or misidentification of objects.
- Inability to capture facial expressions - CLIP may fail to accurately recognize or interpret facial expressions depicted in images, impacting its understanding of human emotions and interactions.

- Lack of attention to details - CLIP may overlook or neglect finer details present in images, leading to incomplete or inaccurate representations of the depicted scene.
- Different activities and positions of people not being encoded properly - CLIP may not effectively encode variations in activities or positions of individuals depicted in images, resulting in misinterpretations of human actions or interactions.
- Failure to account for the size and number of objects/persons/animals present in the image - CLIP may struggle to accurately estimate the size or quantity of objects, persons, or animals within an image, leading to errors in scene understanding.
- Failure to capture the difference in gender roles and activities - CLIP may not adequately capture differences in gender roles or activities depicted in images, potentially resulting in biased or inaccurate interpretations.
- Cultural Misrepresentation - CLIP may misinterpret cultural cues or symbols depicted in images, leading to misunderstandings or misrepresentations of cultural contexts.
- Disregarding object interactions - CLIP may overlook or fail to interpret interactions between objects depicted in images, impacting its understanding of spatial relationships and scene dynamics.
- Different types of transportation/animal confusion - CLIP may struggle to distinguish between different modes of transportation or types of animals depicted in images, leading to confusion or misidentification.

## 5 Conclusion

Our findings not only validate the previously recognized shortcomings of vision models like CLIP, but they also unveil novel systemic faults. These insights emerged from our comprehensive comparisons between CLIP and DINOv2's image embedding capabilities, as well as our in-depth analysis of CLIP's resilience in coping with image transformations. While our investigation was conducted primarily on the Flickr8k dataset, there remains ample potential for future studies to broaden their scope to include other datasets such as ImageNet or BirdSnap. By extending our analysis to diverse datasets, researchers can gain a more holistic understanding of the strengths and weaknesses of vision models across various contexts and domains.